# HalalNet: A Deep Neural Network that Classifies the Halalness Slaughtered Chicken from their Images

*Corresponding author
efahmed@graduate.utm.my

A. Elfakharany*, R. Yusof, N. Ismail, R. Arfa, M. Yunus

Center for Artificial Intelligence & Robotics, Malaysia Japan International Institute of Technology, Universiti Teknologi Malaysia, Jalan Sultan Yahya Petra, 54100 Kuala Lumpur, Malaysia.

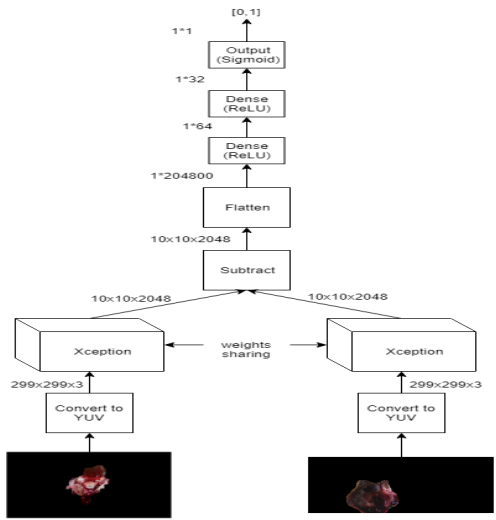

**Abstract**

Halal requirement in food is important for millions of Muslims worldwide especially for meat and chicken products, insuring that slaughter houses adhere to this requirement is a challenging task to do manually. In this paper a method is proposed that uses a camera that takes images of slaughtered chicken on the conveyor in a slaughter house, the images are then analyzed by a deep neural network to classify if the image is of a halal slaughtered chicken or not. However, traditional deep learning models require large amounts of data to train on, which in this case these amounts of data were challenging to collect especially the images of non-halal slaughtered chicken, hence this paper shows how the use of one shot learning [1] and transfer learning [2] can reach high accuracy on the few amounts of data that were available. The architecture used is based on the Siamese neural networks architecture which ranks the similarity between two inputs [3] while using the Xception network [4] as the twin networks. We call it HalalNet. This work was done as part of SYCUT (syriah compliant slaughtering system) which is a monitoring system that monitors the halalness of the slaughtered chicken in a slaughter house. The data used to train and validate HalalNet was collected from the Azain slaughtering site (Semenyih, Selangor, Malaysia) containing images of both halal and non-halal slaughtered chicken

## 1.0 INTRODUCTION

The halal slaughtering process for chicken is achieved by cutting the throat of the chicken through cutting trachea, esophagus and the two carotid arteries and jugular veins without decapitating the head during the slaughtering process [12,13]. To ensure that slaughterhouses adhere to the halal requirements, monitoring of the slaughtering process is required, current methods require inspectors to manually observe and inspect the slaughtered chicken [14]. However, with the rapid increase in slaughtering technology which increases the capacity of slaughter houses can reach up to 3000 chicken/hour, manual inspection is rendered inefficient and time consuming. Hence, an automated system is required and for this reason the SYCUT (syriah compliant slaughtering system) project was initiated.

SYCUT is a monitoring system designed to monitor the slaughtering process of chicken in a slaughterhouse, one of the components of the system is a camera installed on the conveyer on which the slaughtered chicken are hanged, the camera captures images of the cut and an algorithm classifies the cut as halal or non halal, in the case of a non halal cut is detected, an alert system is activated so that the chicken can be removed.

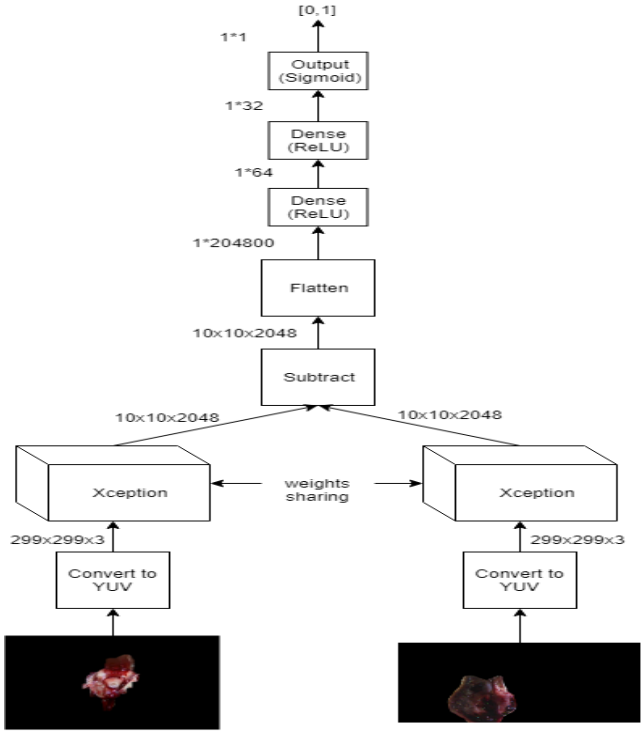

**Figure 1** HalalNet Architecture

Deep learning algorithms are the current state of the art in the field of computer vision [8], in which a neural network with multiple hidden layers and a classification output layer is trained on a labelled dataset. These algorithms have proved their ability to exhibit a good generalization behavior on data that it wasn't trained on which share the same domain of the training data [15]. However, one of it's drawbacks is that it requires a large dataset to train on.

The data collection from the slaughterhouse was challenging as it required disturbing the normal work flow at the site which limited our ability to collect the amount of data sufficient to train a deep neural network to classify the image of the cut if it is halal or non halal. This situation led us into using the one shot learning technique [1] which requires few data per class to train on.

## 2.0 LITERATURE REVIEW

The concept of one shot learning was being explored even before the deep learning popularity rise. In [16], the authors used a Bayesian model to take advantage from knowledge coming from previously learned categories. Probabilistic models represent object categories and a probability density function represent the prior knowledge. While in [17] the authors used RGBD images for gesture recognition, they performed morphological denoising on depth images, then they extracted features based on Extended Motion History Image and fused both features using the Multi-view Spectral Embedding (MSE) algorithm then used Maximum Correlation Coefficient classifier.

The concept of using Siamese networks for one shot learning for image verification was introduced in [3], which consist of twin convolutional networks whose inputs are distinct and their weights are shared, then weighted L1 distance is calculated between the twin feature vectors then passed through a sigmoid activation, the loss function used was cross-entropy following the approach of [10].

In FaceNet [11], a deep convolutional network is used to learn Euclidean embeddings of the images of faces, the network is trained in a manner that the squared L2 distances between embeddings of pictures of the same face is small, while between embeddings of pictures of different faces is large, face verification is done by thresholding the distance between two embeddings.

The matching nets architecture [9,19] uses an external memory to augment the network, while [18,19,20] used an attention network then a classifier.

## 3.0 METHODOLOGY

### 3.1 The Architecture

A deep neural network was used to classify the image of the slaughtered chicken if it is halal or not, the HalalNet architecture was inspired by the Siamese [3] network, the architecture consists of two twin convolutional neural networks sharing their weights, the two networks are fed with two images and each network outputs the features representations of it's input image, the next step is that both features representations are subtracted then passed through a series of fully connected layers then the output layer consists of one sigmoid activated neuron that computes the probability of the two input images are of the same class.

We used the Xception network [4] as each of the twin networks, it is a stack of 36 depthwise separable convolution layers structured into 14 modules with residual connections, with an input shape of (299, 299, 3) and an output shape of (10, 10, 2048), then the output of the twin networks is subtracted then flattened then passed through a 64 neuron fully connected layer with ReLu activation the a 32 neuron layer with ReLU activation then the output layer consists of a single neuron with sigmoid activation to limit the output value between 0 and 1.

The input images to the network is in the YUV color space. The entire architecture is shown in Figure 1. The total number of parameters is around $34e6$ parameters.

### 3.2 Data Preprocessing

The input image was segmented to separate the foreground (the cut in the chicken's neck) from the background using first a (15x15) Gaussian blur filter then the image is converted to the YCbCr color space then Otsu thresholding [6] which results in a binary image that has some holes in the foreground and in the background, to fix this problem closing was used to dilate the holes in the foreground and opening was used to erode the holes in the background as shown in Figure 2. In some cases, the segmentation fails to segment some of the halal images and results in a non-perfect segmentation as in Figure 3 where it shows that the segmentation failed to filter out some of the feathers covered in blood.

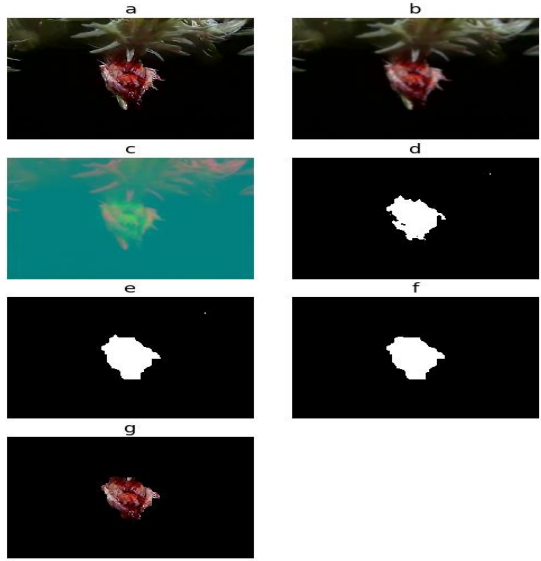

**Figure 2** The image going through the segmentation process. (a) the RGB image acquired from the camera. (b) The blurred image with a (15x15) Gaussian kernel. (c) The YCbCr color space image. (d) The binary image from Otsu thresholding. (e) The image after applying the closing morphological transformation. (f) The image after applying the opening morphological transformation. (g) The final segmented image showing the cut only.

### 3.3 Inference

To perform inference using HalalNet, a set of images with known classes is used to compare the input image against, this set of known images is called control set. The input image is passed through the network paired with a control image, this is done for each class in the control set and the class with higher probability is the class in which the input image is classified.

## 4.0 RESULTS AND DISCUSSION

### 4.1 The Dataset

The dataset used was collected form the Azain slaughtering site (Semenyih, Selangor, Malaysia) containing 737 halal images and 30 non-halal images. To train the network on various cases which can occur in the field, the halal images have different visibility cases for the cut of the slaughtered chicken which are: images of clear and visible cut, images in which the cut is blurred, images in which the cut is covered in blood, images which are dark, image sin which the cut is occluded and images in which the cut is viewed from the side. Samples of these images can be seen in Figure 4 and Table 1 contains the number of images in each case. The halal images where collected by mounting a camera on the conveyor that carries the chicken and capturing images of the cut.

On the other hand, non halal images were hard to collect. The site processes 3000 chicken/hour and to deliberately slaughter some non halal chicken on the conveyer was costly and hard to obtain. In order not to disturb the work flow at the site, the non halal images were collected from chicken slaughtered away from the conveyer leading to the difference in the background between the halal and non halal images as can be seen in Figure 4 and in Figure 5 (a). The segmentation step was taken to lead the neural network to focus on the differences in the cut during training rather than focusing on the backgrounds.

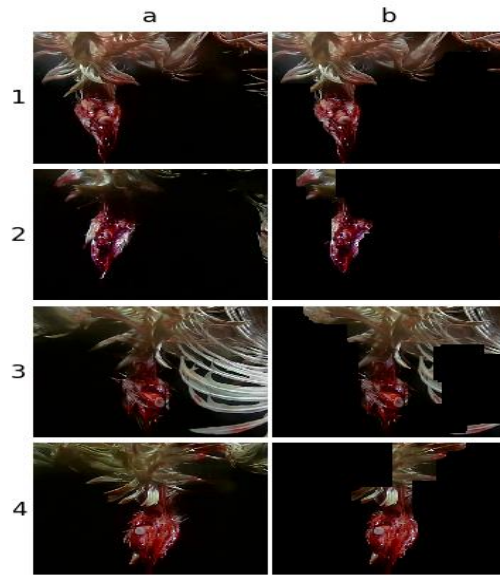

**Figure 3** Samples of failed segmentation on halal images. Column (a) the images before segmentation. Column (b) the images after segmentation.

**Table 1** Halal images cut visibility cases

| Description | No. of images |
| --- | --- |
| Clear visible | 520 |
| Blurred | 13 |
| Bloodied | 126 |
| Dark | 25 |
| Obstructed | 14 |
| Side | 39 |

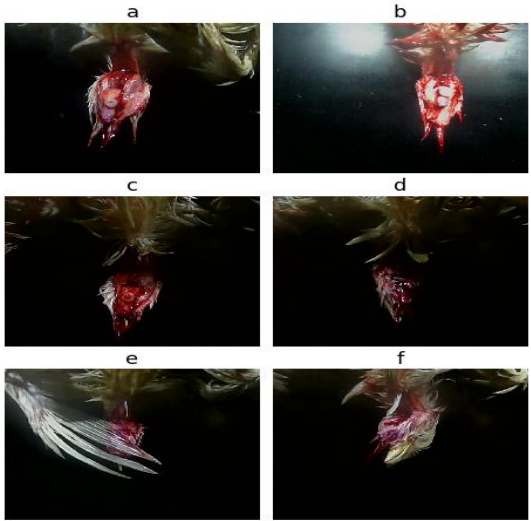

**Figure 4** Samples of the halal dataset. (a) A clear image of the cut. (b) A blurred image of the cut. (c)An image in which the cut is bloodied. (d) A Dark image of the cut. (e) An occluded image of the cut. (f) An image in which the cut is viewed from the side.

**Table 2** Training parameters

| Parameter | Value |
| --- | --- |
| Learning rate | 1e-4 |
| Learning rate decay | 0.99 |
| Optimizer | Adam [5] |
| Batch size | 8 |
| Epochs | 3200 |

### 4.2 Training

To make the network resilient to the segmentation errors and to lead it to focus on the cut, the network was trained on both the segmented and non-segmented data, both kinds of data were stacked into arrays where input data were sampled from, with the probability of sampling the segmented data twice of that of the non-segmented data, the network is trained by randomly selecting pairs of images, when both images are of halal slaughtered chicken or non-halal slaughtered chicken then the label is 1, when one image is of a halal slaughtered chicken and of non-halal then the label is 0.

Data augmentation [8] was used on the input images for multiple reasons. First, to train the network against some changes that can occur while capturing images due to different conditions in the site for example: different lightening conditions or change in the orientation of the image. Second, to artificially increase the amount of data. Third, avoid the problem of overfitting. The augmentation techniques were applied by a probability of 0.5 for each technique, the techniques used were: flip left and right, flip up and down, cropping, padding, scaling, translation, rotation, shearing, wrapping, change of brightness and piecewise affine transformations. Figure 6 shows some of the augmentation techniques.

The dataset was split 70% of the data for training, 15% for cross validation and 15% for final testing.

Both of the twin networks share the same weights and were initialized with Xception's ImageNet weights to transfer and tune the features already learned on the image net dataset while the three fully connected layers were initialized with Xavier uniform initializer [7] and L2 kernel regularization was used, Table 2 includes the parameters used for training.

The loss function used was binary cross-entropy:
$$L = -(y \log(p) + (1 - y) \log(1-p)) \quad (1)$$
Where p is the predicted probability, y is the label, L is the loss

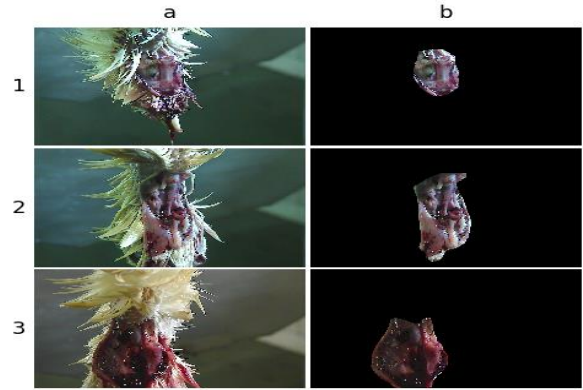

**Figure 5** Samples of the non-halal dataset. Column (a) A full clear image of the non-halal cut. Column (b) A segmented image of the cut.

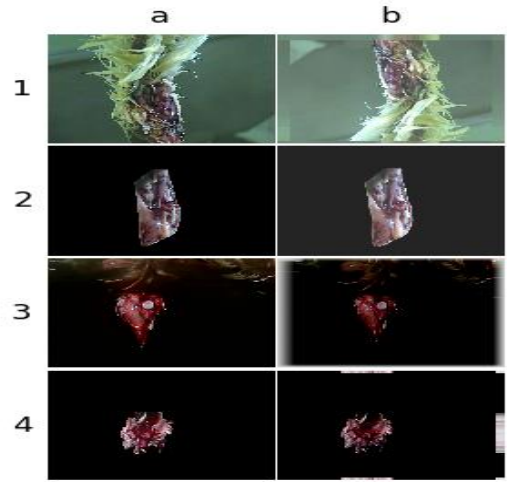

**Figure 6** Samples of Augmented images. Column (a) Images before augmentation. Column (b) Images after augmentation. Row (1) Image of non-halal slaughtered chicken, the augmentation techniques used are: flip left and right, flip up and down, wrapping and increase of brightness. Row (2) Segmented image of non-halal slaughtered chicken, `the augmentation technique used is: increase of brightness. Row (3) Image of a halal slaughtered chicken, the augmentation techniques used are: decrease of brightness, scaling and padding. Row (4) Segmented image of a halal slaughtered chicken, the augmentation techniques used are: decrease of brightness and wrapping.

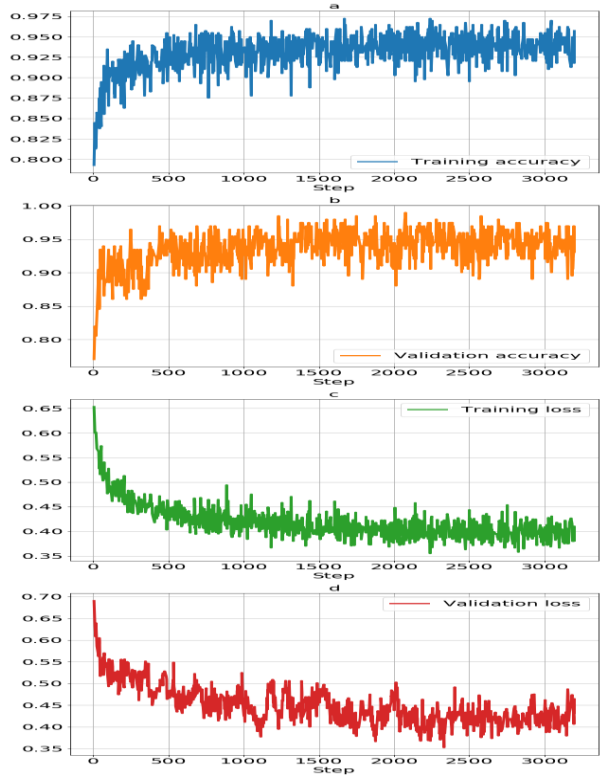

| Metric | Value |
| --- | --- |
| Test Accuracy | 0.9648 |
| Test loss | 0.4136 |
| Precision | 0.9656 |
| Recall | 0.96484 |
| F1 score | 0.96483 |
| True negative | 126 |
| False positive | 2 |
| False negative | 7 |
| True positive | 121 |

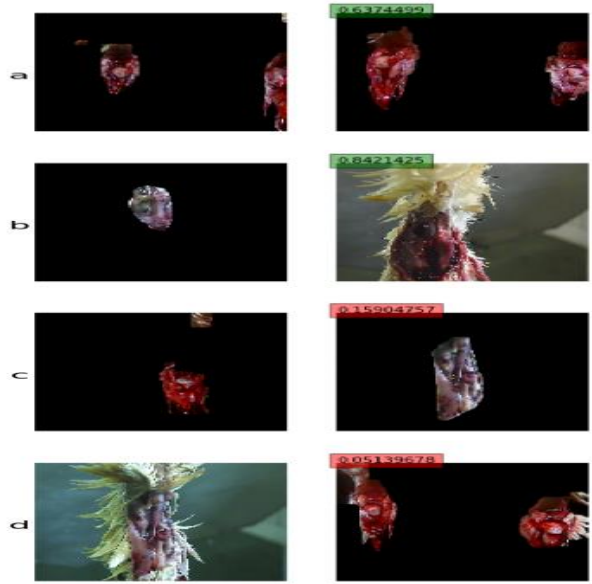

**Figure 7** (a) Training accuracy. (b)Validation accuracy. (c) Training loss: binary cross entropy. (d) Validation loss: binary cross entropy.

### 4.3 Training Results

After training the neural network on the dataset, the accuracy at the end of the training reached 0.9575, accuracy on validation data reached 0.96 as shown in table 3. Figure 7 shows the values of the losses and accuracies after each training epoch.

**Table 3** Training metrics

| Metric | Value |
| --- | --- |
| Training accuracy | 0.9575 |
| Training loss | 0.3809 |
| Validation accuracy | 0.9600 |
| Validation loss | 0.4308 |

To further analyze the performance of the neural network, a batch of 256 pairs of images from the test data set was used to calculate more performance metrics of the neural network, the results are shown in Table 4 which shows that when the neural network misclassifies, it tends to misclassify a pair of images from the same class as a pair of images from different classes more than it tends to misclassify a pair of images from different classes as a pair of images from the same class. Figure 8 shows samples of pairs of images that the network predicted correctly and Figure 9 shows samples of pairs of images that the network predicted incorrectly.

**Table 4** Testing metrics on a batch of 256 pairs of images

**Figure 8** Samples of images where the network predictions were correct (a) a pair of halal images that the network predicted the probability of them being from the same class is 0.637 while images in (b) are both non-halal with predicted probability of 0.8421, in (c) and(d) one of the images is halal and the other is non halal and the predicted probabilities were 0.159 and 0.051 respectively.

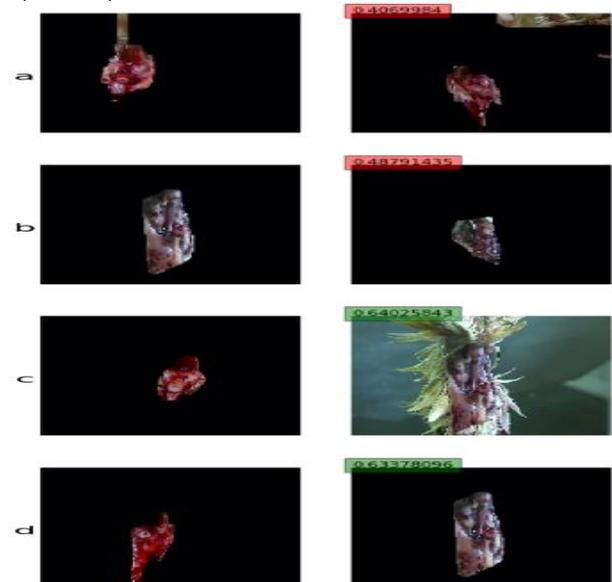

**Figure 9** Samples of images where the network predictions were incorrect (a) a pair of halal images that the network predicted the probability of them being from the same class is 0.407 while images in (b) are both non-halal with predicted probability of 0.488, in (c) and(d) one of the images is halal and the other is non halal and the predicted probabilities were 0.64 and 0.634 respectively.

## 5.0 CONCLUSION

In this paper, we showed that an automatic system that detects the halalness of a slaughtered chicken based on computer vision and deep learning is possible. A Siamese network was trained on few data and achieved approximately 0.95 accuracy. The network performed with high accuracy on data that was properly segmented and data that wasn't segmented and images that were augmented using multiple augmentation techniques, making the network resilient to segmentation errors and any some changes in the environment in which the images are captured.